\documentclass[letterpaper, 10 pt, conference]{ieeeconf}  

\IEEEoverridecommandlockouts                              

\usepackage{cancel}

\usepackage{graphicx}
\usepackage[font=small]{caption}
\usepackage{subcaption}
\usepackage{hyperref}
\usepackage{color}
\usepackage{gensymb}
\usepackage{amsmath}
\usepackage{dsfont}
\usepackage{cite}
\usepackage{adjustbox}
\usepackage{mathtools}
\usepackage{float}
\usepackage{flushend}

\newcommand{\bel}[0]{\ensuremath{\mathrm{Bel}}}

\title{\LARGE \bf   
Probabilistic Object Maps for Long-Term Robot Localization
}

\usepackage[dvipsnames]{xcolor}
\usepackage{booktabs}
\usepackage{multirow}
\usepackage[normalem]{ulem}

\author{Amanda Adkins \and Taijing Chen \and Joydeep Biswas
\thanks{\modified{The authors}\hide{Amanda Adkins, Taijing Chen, and Joydeep Biswas} are with the Department of Computer Science, The University of Texas at Austin, Austin, TX. Email:
        {\tt\small \{aaadkins, taijing, joydeepb\}@cs.utexas.edu }}
\thanks{\modified{This material is based upon work supported by the National Science Foundation (Graduate Research Fellowship Program under Grant No. DGE-2137420, CAREER-2046955), ARO (W911NF-19-2-0333), and Amazon. Any opinions, findings, and conclusions or recommendations expressed in this material are those of the author(s) and do not necessarily reflect the views of the National Science Foundation.}}
}

\newcommand{\AAAcritical}[1]{}
\newcommand{\AAA}[1]{}
\newcommand{\RA}[1]{}
\newcommand{\SM}[1]{}
\newcommand{\jb}[1]{}

\newcommand{\modified}[1]{#1}

\newcommand{\hide}[1]{}

\newcommand{\vs}{\textit{vs.}}

\newsavebox{\twosubbox}

\begin{document}

\maxdeadcycles=500

\maketitle
\thispagestyle{empty}
\pagestyle{empty}

\begin{abstract}
Robots deployed in settings such as warehouses and
parking lots must cope with frequent and substantial changes when localizing in
their environments. While many previous localization and mapping algorithms have explored methods of
identifying and focusing on long-term features to handle change in such environments, we propose a different approach -- can a robot understand the
\emph{distribution} of movable objects and relate it to observations of such
objects to reason about global localization? In this paper, we present
probabilistic object maps (POMs), which represent the distributions of
movable objects using pose-likelihood sample pairs derived from prior trajectories through the environment and use a Gaussian process classifier to generate the likelihood of an object at a query pose. We also
introduce POM-Localization, which uses an observation model based on POMs to perform inference on a factor
graph for globally consistent long-term localization. We present empirical
results showing that POM-Localization is indeed effective at producing globally
consistent localization estimates in challenging real-world environments and that POM-Localization improves trajectory estimates even when the POM is formed from partially incorrect data. 

\end{abstract}

\section{Introduction}

Mobile robots deployed in real world  environments with humans frequently encounter changes due to movable objects. 
Since localization algorithms that assume the world is static fare poorly in such dynamic environments, state-of-the-art long-term localization approaches explicitly model movable and moving objects in an attempt to improve robustness, but come with several limitations. Some of these approaches \cite{zhao2019robust, walcott2012dynamic,  rosen2016towards,nobre2018online, biswas2016episodic} attempt to discover which features persist over long time scales and either discard the remaining features or keep them only for short-term use. This reduces the information available for localization, particularly when such movable objects comprise a large portion of the scene, and can cause the localization estimate to drift over time. Other methods \cite{stachniss2005mobile, krajnik2017fremen, tipaldi2013lifelong} impose assumptions about the configurations or movement patterns of objects that may not match the true 
dynamics of the environment.

\begin{figure}[tbp]
\begin{subfigure}[b]{\columnwidth}
\centering

\includegraphics[width=0.8\columnwidth]{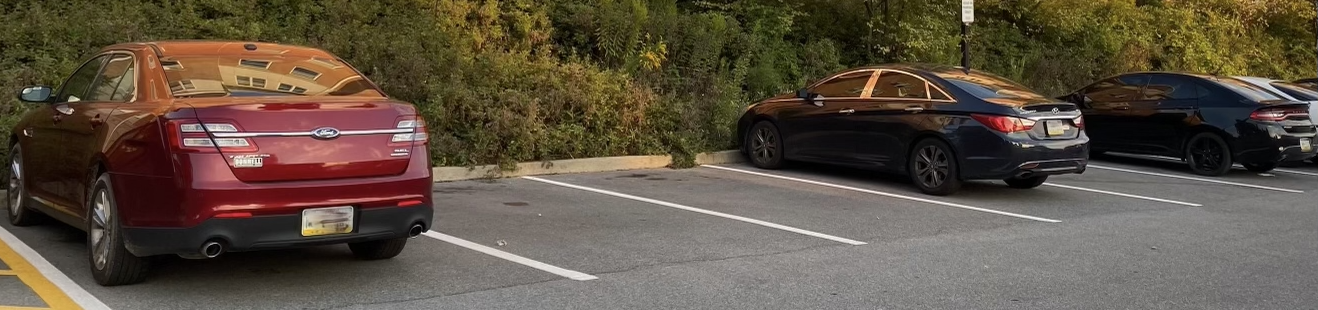}
\end{subfigure}
\begin{subfigure}[b]{\columnwidth}
\centering

\includegraphics[width=0.8\columnwidth]{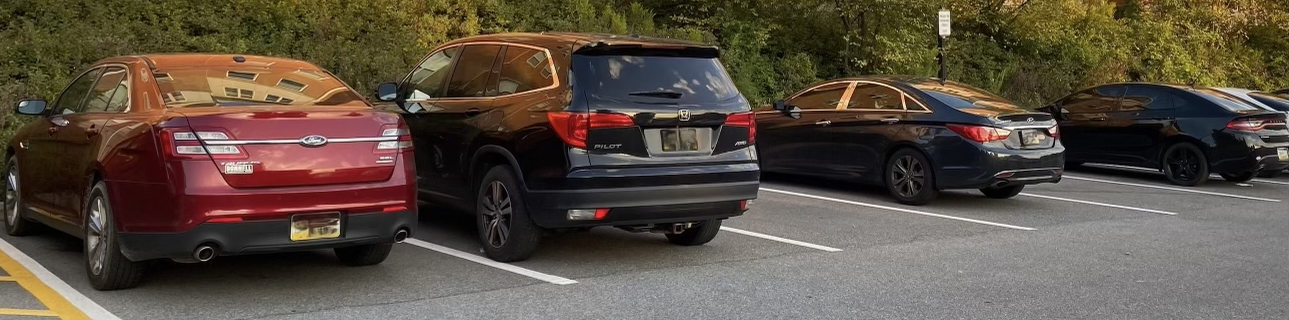}
\end{subfigure}
\begin{subfigure}[b]{\columnwidth}
\centering

\includegraphics[width=0.8\columnwidth]{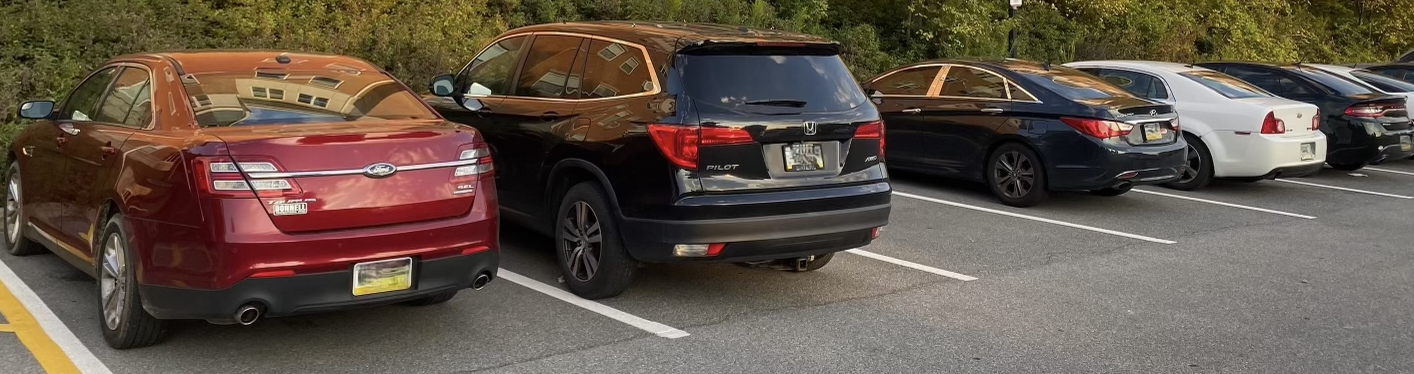}
\subcaption{Parking lot at three different times.}
\label{fig:parking_lot_changes}
\end{subfigure}
\begin{subfigure}[b]{\columnwidth}
\centering

\includegraphics[width=0.65\columnwidth]{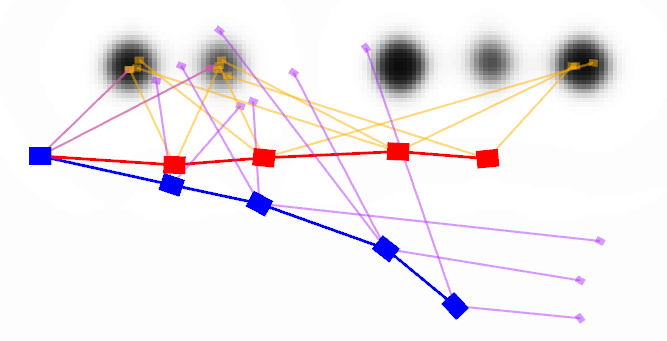}
\subcaption{POM for the above parking configuration, with more frequently occupied spots resulting in higher map values. An initial trajectory is shown in blue with\hide{\modified{ relative}} object detections in purple. \hide{An optimized trajectory from POM-Localization is shown in red. Corresponding object detections \modified{(relative to the optimized trajectory)} in orange line up with peaks in the POM.}
\modified{POM-Localization optimizes the trajectory (shown in red) so the corresponding object detections (shown in orange) are consistent with regions of high probability in the POM.}}
\label{fig:pom_example}
\end{subfigure}
\caption{POM-Localization as applied to a parking lot.}
\label{fig:teaser_fig}
\vspace{-1.5em}
\end{figure}

In many environments, movable objects tend to follow patterns, with some areas more likely to contain objects than others. 
Consider a parking lot like the one in Fig. \ref{fig:parking_lot_changes}: 
the 
scene contains many movable objects, and consequently,
this environment would challenge many localization algorithms. There is no fixed set of configurations and the individual objects may not have consistent periodic movement.  However, such objects do follow a distribution that describes where they are likely to occur and where they are less likely. This holds true for other scenarios, such as pallets and boxes in warehouses or furniture in home and office environments.

We posit that localization algorithms can utilize the distribution of movable objects to improve robustness in real world environments. We \modified{introduce the concept of}\hide{propose a method for creating a} probabilistic object map\modified{s} (POM\modified{s}) \hide{that}\modified{to} model\hide{s} the likelihood of a movable object of a given semantic class occurring at a given pose \modified{and propose a method for POM representation}. The POM is formed in a data-driven manner from object detections from past trajectories and uses Gaussian process classification to generate likelihoods for new query poses. We also introduce the POM-Localization algorithm, which uses POMs to inform localization. The goal of POM-Localization is to prevent significant drift in environments with a high density of movable objects, rather than to improve average-case performance. Given the current estimate for the robot's trajectory and the object detections from that trajectory, POM-Localization calculates where objects would be in the global frame and adds a cost based on the likelihood of an object occurring at the pose, penalizing the trajectory when objects would occur \modified{at}\hide{in} unlikely \hide{positions}\modified{poses}. As shown in Fig. \ref{fig:pom_example}, this formulation is used to optimize the trajectory, resulting in a sequence of poses that best aligns with current object detections, odometry measurements, and the POMs. We also provide a method for incrementally updating \hide{the}\modified{a} POM given a newly optimized trajectory with object detections.
We present experimental results on two different datasets to highlight the performance of POM-Localization in changing environments and to demonstrate the impact of our approach given limited knowledge of the object distribution represented in the POM.

\section{Related Work}
\subsection{Semantic SLAM and Localization}
Similar to our method, semantic landmark SLAM and localization approaches rely on high-level semantic objects rather than low-level features. Many approaches, like \cite{yang2019cubeslam},
require data association by the feature extractor
to associate measurements to landmarks.
This can be challenging, particularly in dynamic environments, and mistakes can degrade results. Some methods try to improve robustness by shifting responsibility for data association to the optimization, allowing correspondence decisions to be updated as more information is obtained. Bowman\hide{,} et al. \cite{bowman2017probabilistic} propose a semantic SLAM approach that integrates data association into the optimization using expectation maximization, while \cite{doherty2019multimodal} and \cite{doherty2020probabilistic} use factor graphs with novel factors that handle data association. Like these approaches, our method uses high-level semantic objects and does not require the feature extractor to resolve data associations. However, our approach avoids the correspondence problem altogether by considering a distribution of objects rather than a discrete set of landmarks. These approaches also differ from ours in that they do not explicitly model changes in environments.

\subsection{Localization and Mapping in Changing Environments}
While many SLAM and localization approaches assume objects in the scene are static, some methods explicitly model moving and movable objects. Some
simply filter movable and/or moving objects from the data \cite{zhao2019robust}. This improves robustness, but results in loss of information that could be valuable for localization. 
Others aim to understand which features persist over long time scales, with remaining features either discarded or kept only for short-term processing. \cite{walcott2012dynamic} removes old data that conflict with more recent information. \cite{rosen2016towards} probabilistically models feature persistence and \cite{nobre2018online} extends this approach to consider relationships between features. Episodic non-Markov Localization (EnML) \cite{biswas2016episodic} matches long-term features to a map and uses short-term features for relative corrections.
Such methods require accurate data association to ensure the correct features are discarded and do not obtain global understanding from short-term features. In scenes with few long-term features, this can lead to drift in the localization estimate. Other approaches impose assumptions about the patterns of movable objects. Such assumptions include a limited range of possible configurations for regions of the space \cite{stachniss2005mobile} or that object movement conforms to some periodicity \cite{krajnik2017fremen} or transition function \cite{tipaldi2013lifelong}. For scenarios such as warehouses, parking lots, or areas with movable furniture, these assumptions  may not appropriately model the real dynamics of the \hide{world}\modified{environment}, and consequently, localization performance could degrade.
        
\subsection{Continuous Mapping}
Appropriate map representations are critical to localization and SLAM. Occupancy grids are commonly used, but introduce errors from discretization. The fixed resolution of occupancy grids is also poorly suited to variable density data and continuous optimization techniques used in many modern localization approaches. Recent works have explored continuous map representations to avoid these shortcomings. Hilbert maps \cite{ramos2016hilbert} model the occupancy of an environment by projecting data into a Hilbert space, while \cite{doherty3dbayesian} 
uses Bayesian Generalized kernel inference.
In \cite{o2012gaussian},
occupancy is estimated with 
Gaussian process classification, 
similar to our approach to modeling object 
likelihood.
These methods have been extended to handle dynamic environments in \cite{senanayake2016spatio} and \cite{ocallaghan2016gaussian}, but 
focus
on observed motion, while the proposed method also accommodates objects that move between robot deployments. 
Further, all of these works focus on raw sensor data and do not incorporate semantics.

\section{Mapping and Localization with\\ Probabilistic Object Maps}

\hide{We introduce the formulation for probabilistic object maps (POMs) and the POM-Localization algorithm that uses these to inform robot localization in environments with movable objects.}
\modified{We introduce the concept of probabilistic object maps (POMs), a POM representation based on Gaussian process classification (GPC), and the POM-Localization algorithm that uses POMs to inform robot localization in environments with movable objects.}
We also outline techniques to increase the speed of POM evaluation and POM-Localization. 
Our approach requires a set of initial trajectories with object detections from which to bootstrap the POMs for each environment. 
The POM-Localization algorithm also requires odometry estimates from either wheel encoders, inertial measurements, or visual odometry.\AAAcritical{Should the following sentence be kept?}
\modified{To avoid assumptions about periodicity, our POM representation does not model temporal patterns\hide{include any temporal components}\AAA{is there a better way to say this?}.}

\subsection{POM Evaluation using Gaussian Process Classification}

The goal of a \modified{POM}\hide{probabilistic object map}\AAAcritical{Should this be abbreviated?} is to estimate the likelihood that an object occurs at a given pose $o^*$. For this, we utilize a variant of\hide{ Gaussian process
classification}\AAA{Can we just use GPC here and not spell it out?} \hide{(}GPC\hide{)}~\cite{bishop2006pattern},
chosen 
for its data-driven nature and continuous differentiability.
Further, this data-driven method allows the POM to capture the distribution of errors arising from observation noise. The POM  estimates the distribution $p(c^*{=}1|o^*)$, where $c^* \in \{0, 1\}$ is a class label indicating whether the pose is occupied by an object, with $p(c^*{=}1|o^*)$ evaluating to 1 when there should always be an object at pose $o^*$ and 0 when there is never an object at the pose.

Let $o_{1:M}$ be the $M$ sample inputs (object poses) for which we have corresponding output values (object occurrence likelihoods).
GPC is closely related to Gaussian process regression (GPR), with the primary difference that the output range of GPR is unbounded, while the output range of GPC is \hide{bound to }$[0, 1]$.  Thus, 
GPC is a more appropriate function approximator for a classification distribution.
GPC transforms the output $a^* \in (-\infty, \infty)$ of GPR using an activation function, such as the logistic function, 
$s(x) = \frac{1}{1+e^{-x}}$. 
For a 3D query pose $o^*$, this transformation via GPR
and GPC is: 
\begin{align}
   &o^* \in \mathds{SE}(3) 
\xrightarrow{\text{GPR}}
 a \in (-\infty, \infty) 
 \xrightarrow{s}
 t \in [0, 1] \\
\equiv \quad
&o^* \in \mathds{SE}(3)  \xrightarrow{\text{GPC}} t \in [0, 1].
\end{align}
Traditionally, GPC assumes training output values $t_{1:M} \in \{0,1\}$ and requires approximations to map these to $a_{1:M} \in (-\infty,
\infty)$ as used by the underlying GPR model. We instead assume that we directly obtain sample values $a_{1:M}$ that are used with GPR and detail this process in Section \ref{section:building_pom}.
The POM estimate $p(c^*{=} 1|o^*)$ is thus given by GPC as 
\begin{align}
    p(c^*{=} 1 | o^*) &= \int p(c^*{=}1 | a^*) p(a^* | a_{1:M}, o_{1:M}, o^*) da^*.
\end{align}
Based on GPR, $p(a^* | a_{1:M}, o_{1:M}, o^*)$ is a normal distribution having mean $\mu$ and variance $\sigma^2$, with $\mu$ given by 
\begin{align}
    \mu &= \mu_0 + K_x^T K_D^{-1}(a-\mu_0), 
\end{align}
where $\mu_0$ is a prior mean \hide{on the interval}\modified{in the range}
$(-\infty, \infty)$, $a-\mu_0$ is a vector of values $a_{1:M}$ less $\mu_0$,
\begin{align}
    K_D &= \begin{bmatrix} 
    k(o_1, o_1) & \dots & k(o_1, o_M) \\
    \vdots & k(o_i, o_j) & \vdots \\
    k(o_M, o_1) & \dots & k(o_M, o_M) 
    \end{bmatrix}, \\
    K_x &= \begin{bmatrix} 
    k(o_1, o^*) \\
    \vdots \\
    k(o_M, o^*)
    \end{bmatrix} ,
\end{align}
and $k(o_i, o_j)$ is a kernel function providing the similarity of $o_i$ and $o_j$. The calculation of $\sigma^2$ is described in Section \ref{section:uncertainty_est}.

$p(c^*{=} 1 | a^*)$ is the logistic function, making $p(c^*{=} 1 | o^*)$ the convolution of a normal distribution and the logistic function. We adapt the approximation from \cite{bishop2006pattern} for such a convolution to incorporate the  prior mean $\mu_0$, giving 
\begin{align}
p(c^*{=}1 | o^*) \approx  s\left(\mu_0 + \frac{\mu-\mu_0}{\sqrt{1+\frac{\pi\sigma^2}{8}}}\right) \label{eqn:gpc_output}.
\end{align}

For POM evaluation, the kernel for computing $\mu$ is a scaled product of a radial basis function (RBF) kernel \cite{bishop2006pattern} on position and a periodic variant of an RBF kernel on orientation. A prior for the likelihood of an object is transformed from $[0, 1]$ by the logit function to obtain $\mu_0$. 

\subsection{Uncertainty Estimation in POMs}
\label{section:uncertainty_est}
Traditionally, GPR assumes that the output for a fixed input is drawn from a normal distribution and the output variance reflects the consistency of the samples near a given input. However, in our case, consistent values are not expected for each query input, so the traditional variance provided by GPR would yield high variance even where there are many samples and confidence in the distribution should be high. Instead, predicting the likelihood of an object occurrence is closer to predicting coin bias, where we have a series of trials (past observations for poses) and we want to know the likelihood of an event (object occurrence at the given pose). Hence, we are computing $\sigma^2$ using an approach derived from the desired properties of the variance: more sample inputs should result in lower variance and samples that are closer to the query pose $o^*$ should result in lower variance than the same number of more distant samples. Consequently, we estimate $\sigma^2$ using the reciprocal of an unnormalized kernel density estimator (KDE) \cite{bishop2006pattern}, which satisfies these conditions. Using this, $\sigma^2$ is given by 
\begin{align}
    \sigma^2(o^*) &= \frac{1}{\mathrm{KDE}(o^*)} = \frac{1}{\sum_{i=1}^M k_{\sigma}(o_i, o^*)}, \label{eqn:sigma_approx}
\end{align}
where $k_\sigma$ has the same form as the kernel for computing $\mu$. 

\subsection{Building Probabilistic Object Maps}
\label{section:building_pom}

To evaluate the likelihood of an object occurring at a given pose using GPC, we need a set of samples $\langle o_i, a_i\rangle_{i=1:M}$, where $o_i$ is a pose in the global frame and $a_i \in (-\infty, \infty)$ represents the object likelihood at the pose based on past trajectories. When the POM is initially created, we use a set of registered trajectories and their object detections. To capture where objects are both likely and unlikely, we obtain $\{o_i\}$ from 1) observed object poses from the registered trajectories as well as 2) poses where objects were not observed, generated by 
sampling from the observed free space around the robot.
The next step is generating values $a_i$ for each sample pose $o_i$. Given a set of object detection poses $S_t=\{s_{t_j}\}$ relative to the robot at time $t_j$ and corresponding variances  $\{\sigma_{t_j}^2\}$, the value $\hat{a}_i$ for a sample pose $\hat{o}_i$ relative to the robot can be obtained using 
\begin{align}
   \hat{a}_i = \max_{s_{t_j}\in S_t}  \mathcal{N}(\hat{o}_i | s_{t_j}, \sigma_{t_j}^2).
\end{align}
If there are no object detections, then $\hat{a}_i$ is 0. We remap each $\hat{a}_i$ from $[0,\infty)$ to $a_i \in (-\infty,\infty)$ using a domain remapping to match the expected input range of the logi\hide{t}\modified{stic} function.\footnote{The choice of the remapping function is not crucial, as long as it is a monotonic injective mapping. 
We use $a_i{=}\log(1 - (1-\exp(\hat{a}_i))^{-1})$.
}
In our approach, a separate POM is created for each movable object class (e.g. separate POMs for cars vs. bicycles).

\subsection{POM-Localization}
POM-Localization incorporates odometry and object detections to estimate the belief over the robot's trajectory. The POM value \hide{associated at}\modified{for} observed object poses is used to calculate the observation likelihood.
We introduce a POM observation likelihood (POM-OL) that is general to several forms of semantic object detections, such as those provided by object pose detectors \cite{ObjDetSurvey} and semantic segmentation \cite{SemanticSegmentationSurvey}. POM-OL requires the ability to draw object pose samples $\hat{o}_{{i_k}_s}\sim p(\hat{o}_{i_k}|r_{i_k})$, where $r_{i_k}$ is relative object detection information.
We first present the POM-Localization algorithm, followed by two such formulations of the POM-OL in Section \ref{section:obs_models}.

The belief over the robot poses $x_{1:n}$ is given by
\begin{align}
\bel(x_{1:n}) &= p(x_{1:n}|x_0, s_{1:n}, u_{1:n}) \\
&\propto \prod_{i=1}^n \prod_{k=1}^{N_i}p(s_{i_k}|x_i)\prod_{j=0}^{n-1} p(x_{j+1}|x_j, u_{j+1}), \label{eqn:bel_init}
\end{align}
where $N_i$ is the number of detections of objects at pose $x_i$, $s_{i_k}$ is the $k$th object detection ($k\in[1,N_i]$) relative to the robot at pose $x_i$, and $u_i$ is the odometry measurement from $x_{i-1}$ to $x_i$ with covariance $\Sigma_\text{odom}$. $s_{i_k}$ is composed of information about the object's pose $r_{i_k}$ and classification variable $c_{i_k}$, and, as we do not consider negative information about objects, $c_{i_k}{=}1$ for all $s_{i_k}$. Next, the observation likelihood $p(s_{i_k} | x_i)=p(r_{i_k}, c_{i_k}{=}1 | x_i)$ can be written as
\begin{align}
    p(s_{i_k} | x_i) = p(r_{i_k}  | x_i, c_{i_k}{=}1)p(c_{i_k}{=}1|x_i)\modified{.}  \label{eqn:obs_like}
    \end{align}
To use the object likelihood in the belief, we marginalize over the true object pose $o_{i_k}$ in the global frame. We can express $p(r_{i_k}  | x_i, c_{i_k}{=}1) $ in terms of $o_{i_k}$ as 
\begin{align}
    &p(r_{i_k}  | x_i, c_{i_k}{=}1){=}\hspace{-4pt}\int_{o_{i_k}}\hspace{-10pt} p(r_{i_k}| o_{i_k}  , x_i) p(o_{i_k} | x_i, c_{i_k}{=}1) do_{i_k} \label{eqn:normal_times_POM}\\
    &=  \int_{o_{i_k}}\hspace{-9pt} p(r_{i_k}| o_{i_k}, x_i) p(c_{i_k}{=}1 | x_i, o_{i_k}) \frac{p(o_{i_k}|x_i)}{p(c_{i_k}{=}1 | x_i)} do_{i_k}. \label{eqn:rik_final}    
    \end{align}
Since the existence of an object is independent of the robot's pose, i.e. $  p(o_{i_k}|x_i) =  p(o_{i_k})$, combining (\ref{eqn:obs_like}) and (\ref{eqn:rik_final}) yields
    \begin{align}
    p(s_{i_k}|x_i) =  \int_{o_{i_k}}\hspace{-9pt} p(r_{i_k} |o_{i_k}, x_i) p(c_{i_k}{=}1 | o_{i_k}) p(o_{i_k}) do_{i_k}. \label{eqn:obs_marg}
\end{align}
Substituting (\ref{eqn:obs_marg}) into (\ref{eqn:bel_init}) gives an updated form for the belief.
We solve for the maximum likelihood estimate $x^*_{1:n}$  by minimizing the negative log likelihood of the belief:
\begin{align}
&{-}\log(\bel(x_{1:n})) \hspace{4pt}\propto\hspace{4pt} \frac{1}{2}\sum_{j=0}^{n-1} ||x_{j+1} \ominus (x_j\oplus u_{j+1})||^2_{\Sigma_\mathrm{odom}} + \nonumber \\
&\sum_{i=1}^n \sum_{k=1}^{N_i} - \log\hspace{-2pt} \int_{o_{i_k}}\hspace{-9pt} p(r_{i_k} | o_{i_k}, x_i)p(c_{i_k}{=}1 |o_{i_k}) p(o_{i_k}) do_{i_k}. \label{eqn:neg_log_likelihood}
\end{align}

The integral above is intractable, so we \hide{estimate}\modified{approximate} it by sampling. 
Noting that the relative pose detection $r_{i_k}$ is independent of $x_i$ conditioned on the relative object pose $\hat{o}_{i_k}=o_{i_k} \ominus x_i $, and applying Bayes' rule, we replace $p(r_{i_k}|o_{i_k}, x_i)$ with $p(\hat{o}_{i_k}|r_{i_k})$ to yield
\begin{align}
\int_{o_{i_k}}\hspace{-9pt} p(r_{i_k} | o_{i_k}, x_i)p(c_{i_k}{=}1 &|o_{i_k}) p(o_{i_k}) do_{i_k} \propto \nonumber \\
\int_{o_{i_k}}\hspace{-9pt} p(\hat{o}_{i_k} &| r_{i_k}) p(c_{i_k}{=}1 |o_{i_k}) p(o_{i_k}) do_{i_k}.
\end{align}
As noted above, we assume that our observation model supports drawing samples from $p(\hat{o}_{i_k} | r_{i_k})$. We can thus approximate the integral using importance sampling by drawing $N_s$ samples $\hat{o}_{{i_k}s}$ from $p(\hat{o}_{{i_k}s} | r_{i_k})$, obtaining the corresponding global frame pose $o_{{i_k}s}$ from the sample using $o_{{i_k}s}{=}x_i \oplus \hat{o}_{{i_k}s}$, and then summing.  With normalization constant $\eta_1$, this gives 
\begin{align}
\int_{o_{i_k}}\hspace{-9pt} p(r_{i_k} | o_{i_k}, x_i)&p(c_{i_k}{=}1 |o_{i_k}) p(o_{i_k}) do_{i_k} \approx \nonumber  \\&\frac{\eta_1}{N_s} \sum_{s=1}^{N_s}  p(c_{{i_k}s}{=}1|o_{{i_k}s})p(o_{{i_k}s}). \label{eqn:object_sample_approx}
\end{align}

We assume $p(o_{i_k})$ is uniform, so we can replace this by a normalization constant $\frac{\eta_2}{\eta_1}$, yielding
\begin{align}
\hspace{-8pt}\frac{1}{N_s}\hspace{-1pt} \sum_{s=1}^{N_s}  p(c_{{i_k}s}{=}1|o_{{i_k}s})p(o_{{i_k}s}){\approx}\frac{\eta_2}{N_s} \hspace{-1pt} \sum_{s=1}^{N_s}  p(c_{{i_k}s}{=}1|o_{{i_k}s}). \label{eqn:eta_factor}
\end{align}

Since the normalization constant is independent of the trajectory, we drop $\eta_2$. Combining (\ref{eqn:neg_log_likelihood}), (\ref{eqn:object_sample_approx}) and (\ref{eqn:eta_factor}) gives
\begin{align}
-\log(\bel(&x_{1:n})) \propto \frac{1}{2}\sum_{j=0}^{n-1} ||x_{j+1} \ominus (x_j\oplus u_{j+1})||^2_{\Sigma_\mathrm{odom}}  + \nonumber\\
\sum_{i=1}^n &\sum_{k=1}^{N_i}- \log \left(\frac{1}{N_s} \sum_{s=1}^{N_s}  p(c_{{i_k}s}{=}1|o_{{i_k}s})\right). \label{eqn:sampled_neg_log_likelihood}
\end{align}

With this formulation and the POM to provide object likelihoods $p(c_{{i_k}s}{=}1|o_{{i_k}s})$, a nonlinear optimizer\modified{, such as \cite{Agarwal_Ceres_Solver_2022},} can be used to find  the trajectory $x_{1:n}$ that best aligns with the movable object observations and odometry.
\subsection{Observation Models}
\label{section:obs_models}
The first observation model uses a relative pose measurement of the object for each observation $r_{i_k}$. For this model, we assume $p(r_{i_k} 
| \hat{o}_{i_k})$ is a normal distribution. Since this is symmetric with respect to $r_{i_k}$ and $\hat{o}_{i_k}$, we can thus sample object poses via the normal distribution. The second observation model uses semantically-labeled points relative to the robot's sensor for each observation $r_{i_k}$. In this model, we approximate $p(\hat{o}_{i_k} | r_{i_k})$ as a histogram over the space of object poses. The value for each bin in the histogram is computed using a method based on the generalized Hough transform \cite{ballard1981generalizing},
enabling
generated samples to be robust to partial observations. We assume a fixed shape and size for each class 
of object. 
Sensor noise is accounted for by adding Gaussian noise to each observed point. 
Given the observed points, we sample points along the object's surface where each of the observed points could occur. From these sampled points, we compute the object's pose in the sensor frame and add a vote for the corresponding bin. To discard outliers, bins with values less than a specified threshold are discarded. The remaining 
candidate poses 
are\hide{ then again} randomly sampled to generate object pose samples to evaluate with the POM.

\subsection{Updating the POM Based on New Trajectories}
We can update the POM to incorporate information from a newly optimized trajectory to better capture the true distribution of objects in the environment. We generate new sample poses and values using the same process outlined in Section \ref{section:building_pom} \modified{with}\hide{using} the sampled object poses based on object observations $\{s_{i_k}\}$, robot fields of view, and optimized poses $x_{1:n}$ from the new trajectory. These new samples are added to the existing samples 
\modified{to form the updated POM, which can be used to optimize}\hide{for the POM for use in optimizing}
subsequent trajectories. 

\subsection{Optimization for Computational Efficiency}

As Gaussian processes are computationally expensive, we employ a number of mechanisms to make the speed of this approach tractable. The first is limiting the samples used to compute $\mu$ and $\sigma^2$ in the POM evaluation to those within a radius of the query pose. We do this by storing the sample poses and values in a KD-tree \cite{bentley1975multidimensional} and constructing a POM at the beginning of each optimization cycle with only samples around the initial estimate for the object pose. Using a subset of the sample poses and outputs when evaluating the POM also reduces the computation time. Let $r_s$ be the fraction of the samples to use and $M$ be the full number of samples. 
We  adjust the POM evaluation by modifying (\ref{eqn:sigma_approx}) to sum instead over $N=r_s M$ random samples and compensate for subsampling by multiplying the full equation by $r_s$.
We can also scale the computational power needed by modifying the number of samples $N_s$ used to approximate the marginalization over object poses. Lastly, using only a subset of the object detections and optimizing over a window of only the most recent nodes in the trajectory improves computation time. 

\section{Experimental Results} 
We present results from two sets of experiments\footnote{\label{code_footnote}The code for POM-Localization and our experiments is available at \url{https://github.com/ut-amrl/pom_localization} and \modified{uses \cite{Agarwal_Ceres_Solver_2022} for optimization}.} that evaluate POM-Localization's ability to 1) accurately estimate trajectories when knowledge of the distribution of movable objects is limited and the correctness of the POM is varied, and 2) ensure consistency in global localization over long time scales in environments with movable objects. We also provide qualitative results for a trajectory in a parking lot using semantically-labeled points. In these experiments, we focus on lidar-based approaches, as the increased field of view and sensor range associated with lidar are important for localization in changing environments.

\subsection{Observation Models}
\begin{figure}[tbp]
	\centering
	\begin{subfigure}[t]{\dimexpr0.29\columnwidth}
		\includegraphics[height=1.8cm]{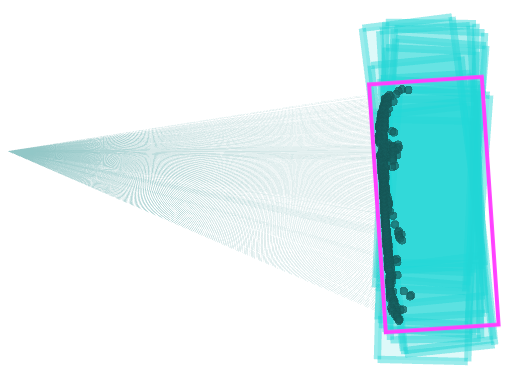}
	\end{subfigure}
	\begin{subfigure}[t]{\dimexpr0.29\columnwidth}
		\includegraphics[height=1.8cm]{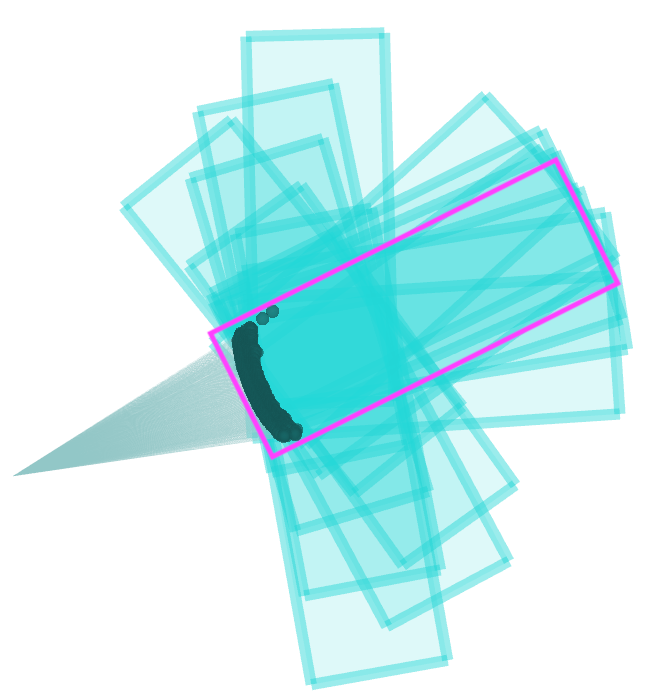}

		\centering
		
	\end{subfigure}
		\begin{subfigure}[t]{\dimexpr0.29\columnwidth}
    \includegraphics[height=1.8cm]{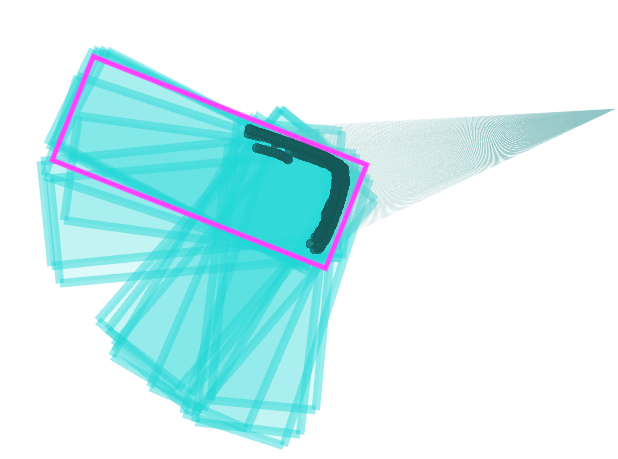}
	\end{subfigure}
    \caption{Examples of observed point clusters and generated object pose samples. True object poses are shown in pink.}
    \label{fig:observed_points_and_samples}
    \vspace{-1.5em}
\end{figure}

These experiments use the two observation models described above. For results using the Hough transform-based model, we approximate cars as 2D rectangles the size of a Toyota Camry. We generate votes for object poses from line segments connecting neighboring points by choosing a random location on the rectangle where the line segment would fall. Examples of observed points and corresponding generated samples are shown in Fig. \ref{fig:observed_points_and_samples}.

\subsection{Accuracy Given Limited Object Distribution Knowledge}

\begin{table*}[tbp]
\centering
\small
\
\begin{tabular}{clrrrrrrrrrr}
& KITTI Sequence Number                       & 00    & 02      & 03   & 04   & 05   & 06   & 07   & 08     & 09   & 10   \\
\toprule
& LeGO-LOAM                              & 27.12 & 1630.55 & 3.51 & 4.11 & 7.97 & 1.19 & 0.76 & 113.44 & 6.48 & 1.76 \\
\hline
\multirow{5}{*}{\shortstack[1]{Object\\ Detections}}
& POM-Localization (0-100) & 29.46 & 1630.55 & 3.51 & 4.11 & 7.97 & 0.80 & 0.76 & 113.44 & 6.48 & 1.76 \\
& POM-Localization (20-80) & 22.93 & 1630.55 & 3.55 & 4.11 & 1.83 & 0.37 & 0.20 & 113.44 & 6.48 & 1.76 \\
& POM-Localization (50-50) & 1.73 & \textbf{1322.19} & 1.39 & 4.06 & 1.95 & 0.08 & \textbf{0.03} & 106.86 & 0.80 & 0.33 \\
& POM-Localization (80-20) & \textbf{1.50} & 1332.74 & 0.89 & {4.10} & 1.71 & 0.09 & 0.04 & 113.42 & 0.81 & 1.06 \\
& POM-Localization (100\textsuperscript{+}-0)  & 1.59  & 1333.48 & \textbf{0.82} & \textbf{4.04} & \textbf{0.18} & \textbf{0.04} & 0.04 & \textbf{97.58}  & \textbf{0.23} & \textbf{0.17} \\
\hline
\multirow{5}{*}{\shortstack[1]{Semantic\\ Segmentation}} & POM-Localization (0-100) & 37.37 & 1553.24 & 3.52 & 4.11 & 5.43 & 1.63 & 0.79 & 111.50 & 9.78 & 3.23 \\
& POM-Localization (20-80) & 6.30 & 1570.66 & 3.22 & 4.09 & \textbf{1.14} & 0.28 & 0.22 & 97.24 & 1.08  & 1.97 \\
& POM-Localization (50-50) & 0.37 & \textbf{1203.31} & 2.00 & \textbf{4.06} & 2.27 & 0.24 & 0.06 & 91.43 & 0.80 & 0.96 \\
& POM-Localization (80-20) & 0.31 & 1408.15 & 0.88 & 4.09 & 1.60 & 0.17 & 0.07 & 69.33 & 0.60 & 0.79 \\
& POM-Localization (100\textsuperscript{+}-0)  & \textbf{0.28} & 1417.72 & \textbf{0.85} & \textbf{4.06} & 1.30 & \textbf{0.15} & \textbf{0.05} & \textbf{36.51}
& \textbf{0.45} & \textbf{0.60}
\end{tabular}
\caption{Absolute Trajectory Error (m) on KITTI dataset sequences for LeGO-LOAM and POM-Localization using observation models for object detections and semantic segmentation.}
\label{fig:KITTI_results}
\vspace{-1.5em}
\end{table*}

We aim to understand how quality of data used to form the POM impacts trajectory estimates when we have not captured the full distribution of objects and we assume low confidence in our map, which would occur when we have not collected sufficient past trajectories to converge to the true distribution. To do so, we measure accuracy using absolute trajectory error (ATE) on 10 sequences from the KITTI dataset \cite{kitti_odometry} and compare against the output of LeGO-LOAM \cite{legoloam2018}, a state-of-the-art lidar-inertial odometry and mapping algorithm. Though our method supports 3D, we use a 2D projection for these experiments\modified{ to match the 2D projection used for object observations}.
The trajectory estimates of LeGO-LOAM serve as our odometry constraints.
We demonstrate our approach using both semantically-labeled points and object poses. In both cases, data is derived from ``car'' instances in the SemanticKITTI dataset \cite{semantic_kitti}, which contains labels for lidar scan points in the KITTI benchmark  (sequence 01 was excluded due to lack of ``car'' observations). The labeled points come directly from the dataset and object detections are created by transforming the global pose of each object into the frame in which the instance was observed.

To assess the impact of the correctness of the POM, we test five different configurations. In all cases, we generate the POM from a single simulated past trajectory, and thus have low confidence in our distribution to emulate the range of performance when bootstrapping the POM from one or few observed past trajectories. The initial step in creating the POM is generating poses where cars occurred in the past trajectory: $X$\% of simulated car poses for the first four configurations are obtained by selecting poses from the current trajectory's observed cars and adding a small amount of Gaussian noise (0.4 m), with the remaining Y\% randomly placed in the environment.
As $X$ increases, the correctness of the POM increases. We chose $X$ from $\left\{0,20,50,80,100\right\}$ in our experiments, and these are denoted ``POM-Localization ($X$-$Y$)'' in Table \ref{fig:KITTI_results}. In the last configuration, POM-Localization  (100\textsuperscript{+}-0), the POM is created from the same car poses that were used to generate the detections without any added noise, simulating perfect knowledge of the distribution and a deterministic environment in which movable objects are always at the same poses. In all cases, once the simulated car poses are selected, we generate the POM following the steps in section \ref{section:building_pom}, with the ground truth trajectory as the prior trajectory, the relative poses of nearby simulated cars with added noise as past object detections, and a fixed-radius region around the trajectory poses as the free space for obtaining off-detection samples. 

Table \ref{fig:KITTI_results} shows the ATE for LeGO-LOAM and our approach with the five configurations using object detections and semantic segmentation. When the POMs are completely misaligned with the observations as in POM-Localization (0-100), estimates are comparable to the LeGO-LOAM results. For the results using object detections, estimates for all but one sequence have the same or lower error and the remaining estimate having only 8.6\% higher ATE than the LeGO-LOAM result. For the semantic segmentation results, estimates for four sequences have somewhat worse error for the completely misaligned POM configuration. However, it is unlikely in practice that all observations in a given trajectory will be completely disjoint from past data. This can also be mitigated by placing higher weight on odometry; however, we aimed to balance the worst-case and best-case performances. As POM correctness increases, results generally improve, with the ATE of most trajectories generated using the POM created from perfect knowledge substantially lower than the ATE for LeGO-LOAM estimates. In most cases, even a half-correct map substantially improves the trajectory estimates. The only sequences for which our approach does not substantially improve upon LeGO-LOAM are 02, 04 and 08. In all of these sequences, there were trajectory segments that simultaneously had substantial odometry error and few or no object detections, and, as odometry is weighted highly relative to object detections, the drift was not corrected and persisted for the remainder of the trajectories.

\subsection{Consistency Over Trajectories in Changing Environments}
\begin{figure}[tbp]
\begin{subfigure}[b]{\columnwidth}
\centering
\includegraphics[width=0.85\columnwidth]{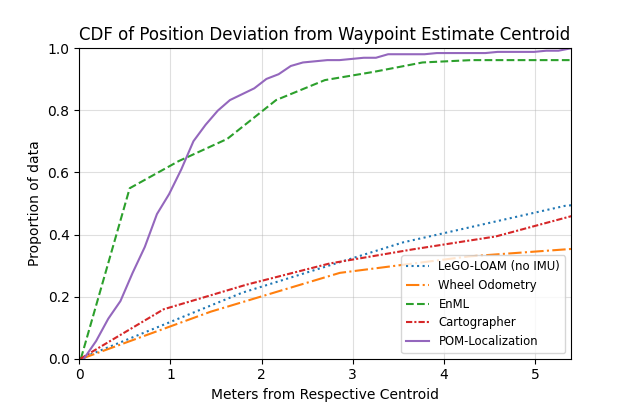}
\subcaption{CDF of position estimate deviation.}
\label{fig:transl_est_deviation_cdf}
\end{subfigure}
\begin{subfigure}[b]{\columnwidth}
\centering
\includegraphics[width=0.85\columnwidth]{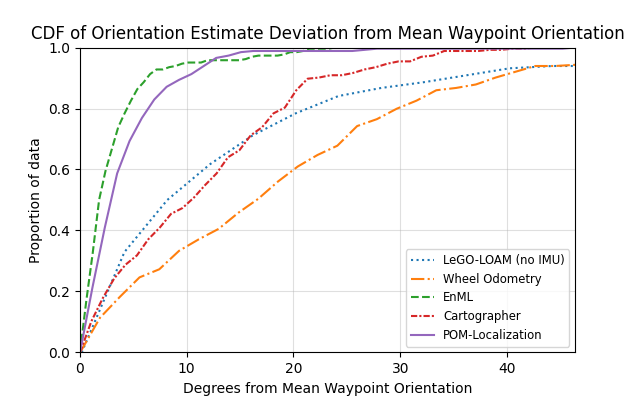}
\subcaption{CDF of orientation estimate deviation.}
\label{fig:orientation_est_deviation_cdf}
\end{subfigure}
\caption{Position and orientation consistency across approaches. An optimal algorithm would quickly rise to 1.}
\label{fig:cdfs}
\vspace{-1.5em}
\end{figure}
\begin{figure*}[t]
	\centering
	\begin{subfigure}[t]{\dimexpr0.29\textwidth}
		\includegraphics[width=\columnwidth]{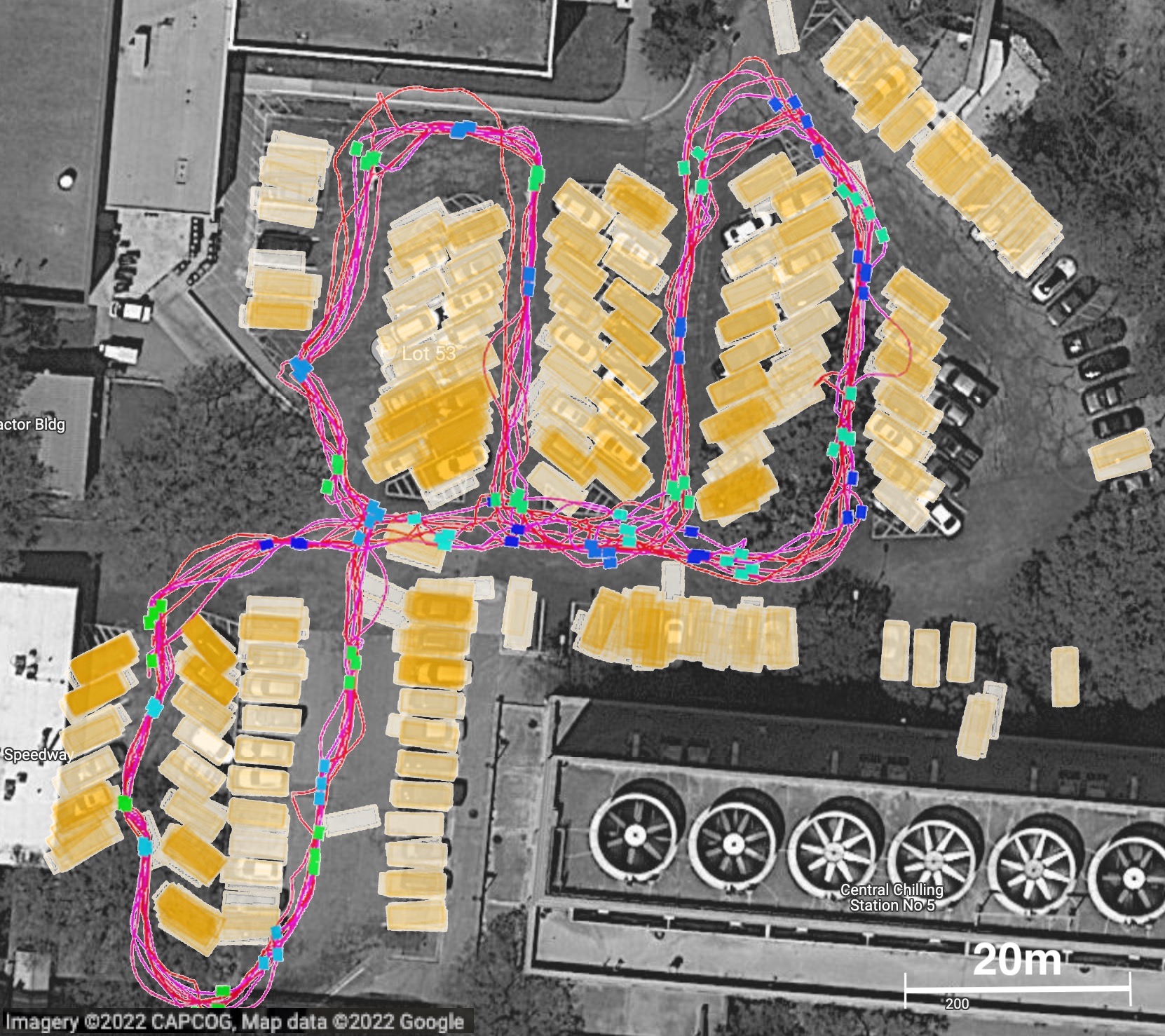}
		\subcaption{POM-Localization with object detections}
		\label{fig:pom_obj_det_parking}
	\end{subfigure}
	\begin{subfigure}[t]{\dimexpr0.19\textwidth}
		\includegraphics[width=\columnwidth]{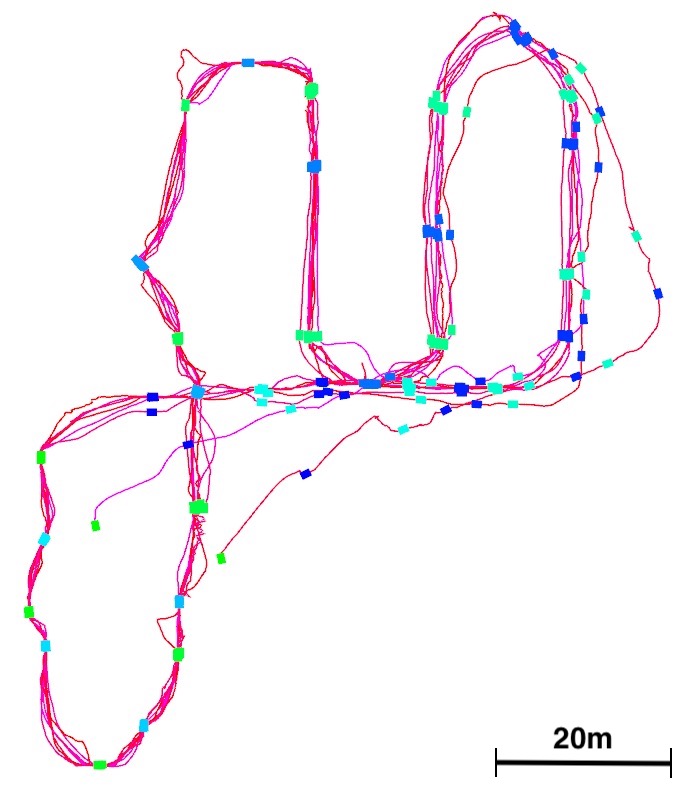}
		\subcaption{EnML}
		\label{fig:enml_parking_lot}
	\end{subfigure}
	\begin{subfigure}[t]{\dimexpr0.21\textwidth}
		\includegraphics[width=\columnwidth]{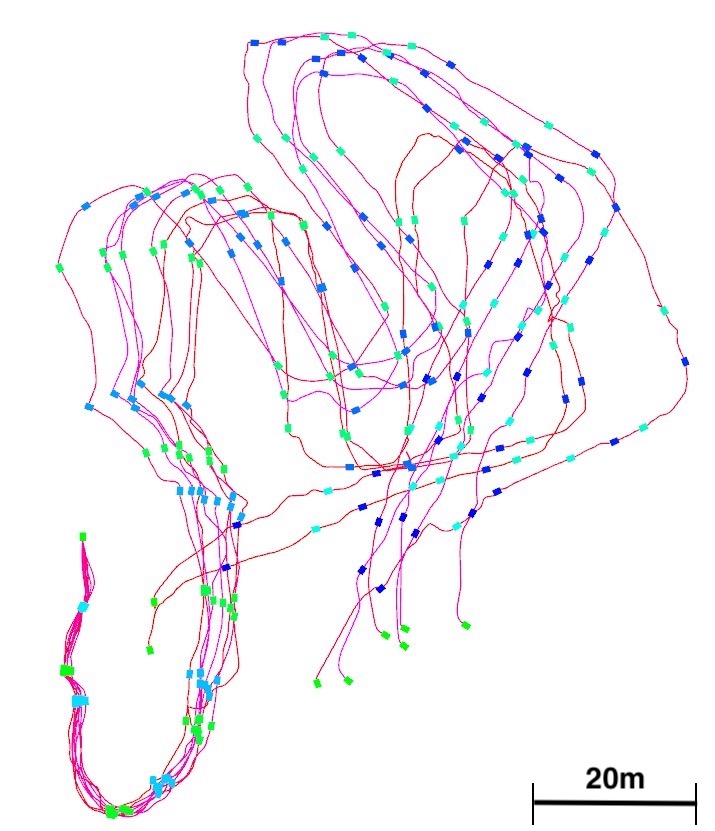}
		\subcaption{Cartographer}
		\label{fig:cartographer_parking_lot}
	\end{subfigure}
	\begin{subfigure}[t]{\dimexpr0.29\textwidth}
		\includegraphics[width=\columnwidth]{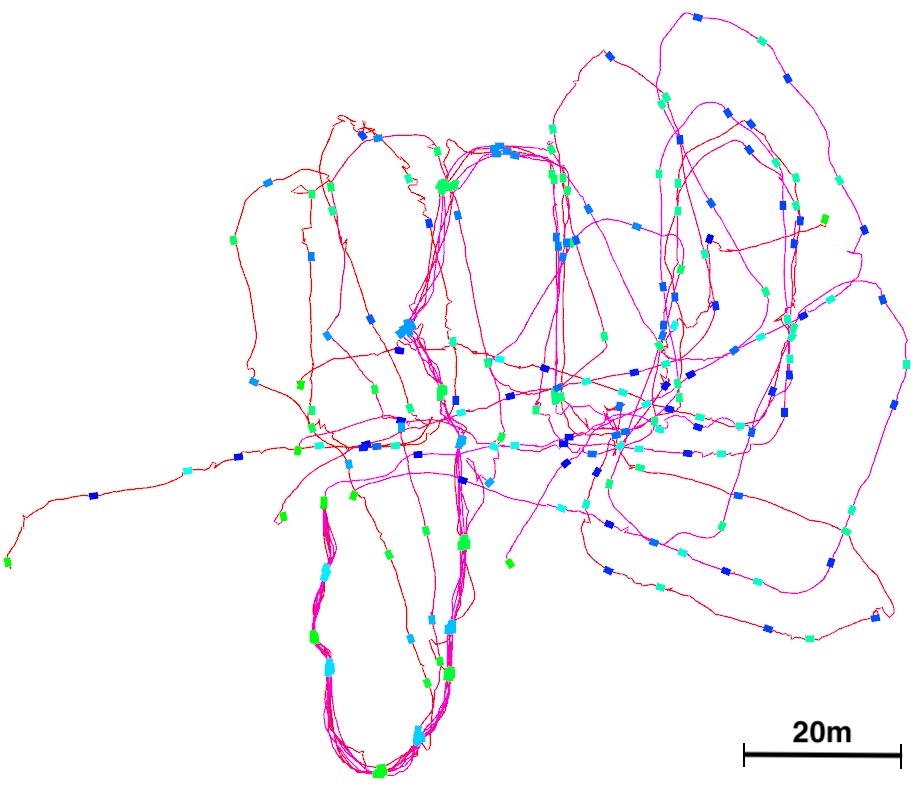}
		\subcaption{LeGO-LOAM}
		\label{fig:lego_loam_parking_lot}
		\centering
		
	\end{subfigure}
	\caption{Plots of trajectories through UT Austin Lot 53 as estimated by the approaches with highlighted blue/green waypoints. Performance of an approach is good when all estimates for a given waypoint are colocated. POM-Localization results are overlayed on a satellite view and shown with aggregated object poses from all trajectories.}
	\label{fig:lot53_traj}
	\vspace{-1em}
\end{figure*}

To understand our approach's ability to produce consistent localization estimates over long time scales in changing environments, we collected \modified{eight }trajectories\hide{ over eight sessions}{\modified{ \hide{across}\modified{over} four days}} in a UT Austin campus parking lot using a Clearpath Jackal with a Velodyne VLP-16 Lidar. In each trajectory, the robot started and ended at the same location and visited 32 waypoints consistent across all trajectories, while the segments between the waypoints varied. 
\modified{The trajectories had an average length of 372 meters and an average duration of 26 minutes. The percent of mapped parking spots occupied by cars ranged from 11\% to 92\%{, with an average of 60\%}. }

Data used for these experiments were collected in 2D. 
Odometry constraints are obtained from wheel odometry and object detections are derived from human-provided annotations of point clouds. The POM is generated from a manually-created map of parking spots\hide{.}\modified{ \hide{This map was manually created }to demonstrate resilience to deviations from the true car distribution and to \hide{remove the variable of}\modified{eliminate} the quality of the localization algorithm used to bootstrap the POM{ as a variable in \hide{our}\modified{the} results}.} We simulate five possible parking configurations by randomly selecting 70\% of the parking spots to be occupied and placing cars according to a normal distribution around each selected spot. Samples to form the POM are created by simulating trajectories through each of the parking configurations, transforming the object poses to the frame of each node, and adding Gaussian noise to simulate noisy detection in past trajectories. Off-detection sample poses are drawn from within a\AAA{consider adding fixed -- not currently space for this} radius around nodes in each simulated trajectory.


We assess performance by measuring consistency of waypoint estimates over the eight trajectories. Position estimate consistency is evaluated by calculating the centroid of all estimates for each waypoint across all trajectories and finding the distance of each estimate from the centroid. Similarly, the 
orientation consistency is measured by computing the mean orientation estimate for each waypoint and the distance of each estimate from the corresponding mean orientation. We show the results of POM-Localization \hide{with}\modified{compared to} estimates from wheel odometry, Cartographer \cite{Cartographer2016} (a commonly used mapping and localization package), LeGO-LOAM \cite{legoloam2018} without use of an inertial measurement unit (IMU), and EnML \cite{biswas2016episodic}.
Cartographer built a map during a single run and used this map in localization mode for the remainder of the trajectories, while EnML matches long-term features to a manually-created map and uses short-term features for local matching. It should be noted that the map and point clouds used by EnML are higher fidelity than the object detections and POM used by POM-Localization.

Fig. \ref{fig:cdfs} shows the cumulative distribution functions (CDFs) for the position and orientation estimate consistency. For both position and orientation, \hide{while }POM-Localization significantly outperforms raw wheel odometry, Cartographer, and LeGO-LOAM\hide{,}\modified{.} EnML has slightly better average-case performance\modified{; t}\hide{. T}his is expected\hide{, given the difference in information sources for the two algorithms}\modified{, as the higher-fidelity data used by EnML enables a higher-precision result}. However, POM-Localization has better or comparable worst-case performance: a greater portion of the orientation estimates are within 15 degrees of the mean for POM-Localization as compared to EnML and no position estimate from POM-Localization deviates more than 5.5 meters from its waypoint centroid. The trajectories and waypoints estimated by POM-Localization, EnML, Cartographer, and LeGO-LOAM are shown in Fig. \ref{fig:lot53_traj}. These plots support the conclusions drawn from Fig. \ref{fig:cdfs}; the estimates from LeGO-LOAM and Cartographer drift substantially after the initial segment of the trajectory, while POM-Localization results in slightly greater spread for the waypoint estimates than EnML, but unlike EnML, has no trajectory estimates that diverge significantly. Fig. \ref{fig:pom_obj_det_parking} also shows that the object pose estimates align with the true parking spots and with each other over time. 

\subsection{Parking Lot Localization With Labeled Points}

\begin{figure}[t!]
\centering
          \includegraphics[width=.7\linewidth]{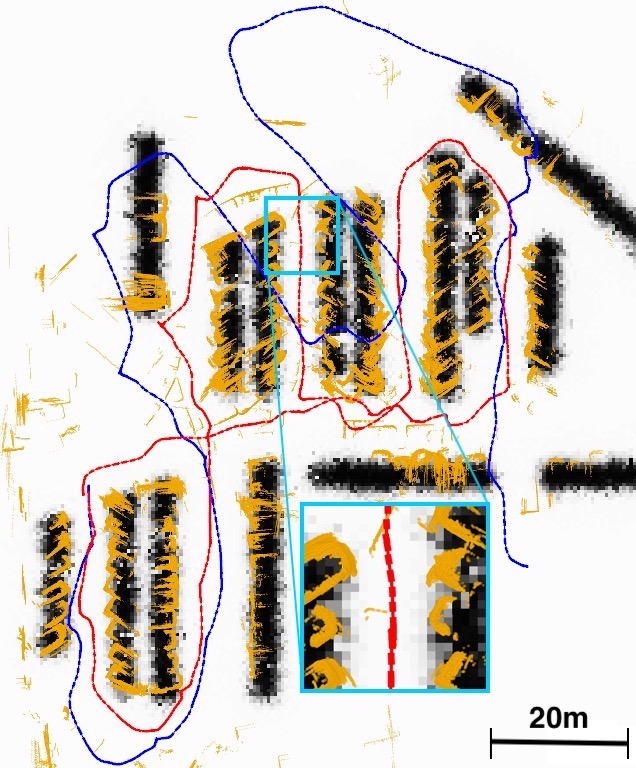}
\caption{POM-Localization optimized trajectory (red) \vs{} odometry estimate (blue), with semantic point observations in orange, and POM in greyscale. Inlay shows alignment of semantic point observations of cars and the alignment with the POM.} 
\label{fig:parking_lot_seg_traject}
\vspace{-1.5em}
\end{figure}

Fig. \ref{fig:parking_lot_seg_traject} shows a trajectory before and after optimization using POM-Localization with semantically-labeled lidar points for car clusters. The trajectory from odometry is shown in blue and the optimized trajectory is shown in red. Object observations for cars are semantically-labeled clusters of points from a 3D point cloud obtained using a heuristic-based segmentation algorithm, which can be found in our repository.$^2$ This algorithm uses PCL \cite{Rusu_ICRA2011_PCL} for clustering and \modified{returns clusters that}\hide{filters clusters that do not} match expected properties of cars. The background of the figure shows a discretized version of the POM, with areas likely to have cars in black and those unlikely to have cars in white. The semantically-labeled points, transformed using the robot pose estimates, are shown in orange. The car segmentation algorithm gives several false positive detections resulting from other objects in the environment and cars in motion are also observed. Despite this, POM-Localization is able to optimize the trajectory so that the stationary cars 
line up with peaks in the POM. 
There is some variation in the resulting pose of each car. However, this is to be expected: the POM is not sharply peaked, so that it can accommodate variations in parked car poses over time, and POM-Localization works without performing data association between sensor data across timesteps. The optimized trajectory begins and ends at approximately the same pose, showing that POM-Localization successfully corrected drift from odometry.


\section{Conclusion and Future Work}
This paper \modified{presents}\hide{introduces} probabilistic object maps\AAA{Should this be switched to the acronym?} to model the distribution of movable objects in an environment. We also introduce POM-Localization to incorporate object detections and corresponding POMs to achieve globally consistent localization in changing environments, thus enhancing the ability of robots to perform autonomously.

There are a number of interesting areas for future exploration. Though our enhancements improved speed, investigation of other approximations may further reduce computation time. 
Future improvements could also include modifying our model to work with monocular visual detectors.\hide{As the POM-Localization factors are modular, they}
\modified{Additionally, the POM-Localization observation factors }could\hide{ also} be integrated into an approach with other factors, such as short-term features or loop closures, allowing the combined approach to gain the benefits of all such factors.
We would also like to explore use of the Generalized Hough transform \cite{ballard1981generalizing} with a more complex shape model or learned method similar to VoteNet \cite{votenet} to propose occulsion-tolerant object pose samples for more complex models. \modified{If the object periodicity is known, time could\hide{ also} be incorporated into the GPC kernel to model temporal patterns. }Finally, POMs could be applied to other problems such as navigation to avoid likely occupied areas or finding objects in home environments.







\bibliographystyle{IEEEtran}
\bibliography{bibliography}

\end{document}